\documentclass[conference]{IEEEtran}
\IEEEoverridecommandlockouts
\makeatother
\usepackage{cite}
\usepackage{amsmath,amssymb,amsfonts}
\usepackage{algorithmic}
\usepackage{graphicx}
\usepackage{textcomp}
\usepackage{booktabs}
\usepackage{xcolor}
\usepackage{float}
\def\BibTeX{{\rm B\kern-.05em{\sc i\kern-.025em b}\kern-.08em
    T\kern-.1667em\lower.7ex\hbox{E}\kern-.125emX}}
\begin{document}

\title{Credit Card Fraud Detection\\ Using Advanced Transformer Model
}

\author{
\IEEEauthorblockN{1\textsuperscript{st*} Chang Yu}
\IEEEauthorblockA{
    \textit{Independent Researcher} \\
    \textit{Northeastern University} \\
    Boston, MA, 02115, USA\\
    Email: chang.yu@northeastern.edu}
\and
\IEEEauthorblockN{2\textsuperscript{nd} Yongshun Xu}
\IEEEauthorblockA{
    \textit{Independent Researcher} \\
    \textit{University of Massachusetts Lowell} \\
    Lowell, MA, 01850, USA\\
    Email: Yongshun\_Xu@student.uml.edu}
\and
\IEEEauthorblockN{2\textsuperscript{nd} Jin Cao}
\IEEEauthorblockA{
    \textit{Independent Researcher} \\
    \textit{Johns Hopkins University}\\
    Baltimore, MD, 21218, USA \\
    Email: caojinscholar@gmail.com}    
\and
\IEEEauthorblockN{2\textsuperscript{nd} Ye Zhang}
\IEEEauthorblockA{
    \textit{Independent Researcher} \\
    \textit{University of Pittsburgh} \\
    Pittsburgh, PA, 15203, USA\\
    Email: yez12@pitt.edu}

 \and
\IEEEauthorblockN{3\textsuperscript{rd} Yixin Jin }
\IEEEauthorblockA{
   \textit{Independent Researcher} \\
\textit{University of Michigan, Ann Arbor}\\
Ann Arbor, MI 48109, USA \\
Email: jyx0621@gmail.com}
\and
\IEEEauthorblockN{3\textsuperscript{rd} Mengran Zhu}
\IEEEauthorblockA{
    \textit{Independent Researcher} \\
    \textit{Miami University} \\
    Oxford, OH, 45056, USA \\
    Email: mengran.zhu0504@gmail.com}
}

\maketitle

\begin{abstract}
With the proliferation of various online and mobile payment systems, credit card fraud has emerged as a significant threat to financial security. This study focuses on innovative applications of the latest Transformer models for more robust and precise fraud detection. To ensure the reliability of the data, we meticulously processed the data sources, balancing the dataset to address the issue of data sparsity significantly. We also selected highly correlated vectors to strengthen the training process. To guarantee the reliability and practicality of the new Transformer model, we conducted performance comparisons with several widely adopted models, including Support Vector Machine (SVM), Random Forest, Neural Network, Logistic Regression, XGBoost, and TabNet. We rigorously compared these models using metrics such as Precision, Recall, F1 Score, and ROC AUC. Through these detailed analyses and comparisons, we present to the readers a highly efficient and powerful anti-fraud mechanism with promising prospects. The results demonstrate that the Transformer model not only excels in traditional applications but also shows great potential in niche areas like fraud detection, offering a substantial advancement in the field.
\end{abstract}


\section{Introduction}
The emergence of information technologies such as mobile internet and big data has led to the rise of internet finance, bringing convenience to people's daily lives. However, along with this convenience comes the phenomenon of internet financial fraud, with credit card transactions being the primary target of various fraudulent activities. The rapid development of internet financial services has significantly increased the probability of credit card fraud incidents, resulting not only in immeasurable losses for individuals and businesses but also in numerous socio-economic problems.

A major technical challenge lies in the fact that fraudulent transactions account for only a small portion of credit card payments, making it challenging for machine learning algorithms to detect such sensitive behavior\cite{ yuan2024research}. As the most powerful and capable model to date, the Transformer large model exhibits revolutionary identification capabilities, making it particularly suitable for recognizing fraudulent data in small samples.

In recent years, various machine learning techniques have been applied to tackle the problem of fraud detection. Neural Networks (NN)\cite{chen2022ba}, for instance, have been widely used due to their ability to learn complex features and relationships from large datasets \cite{lecun2015deep}. NNs can automatically extract features from raw data and create a hierarchical representation, enabling them to detect fraudulent activities with high accuracy \cite{abdallah2016fraud}.

Another popular approach is XGBoost, an ensemble learning method that combines multiple decision trees to create a robust and accurate model \cite{chen2016xgboost, akbulut2023hybrid}. XGBoost has proven to be effective in fraud detection tasks, as it can handle imbalanced datasets and capture non-linear relationships between features\cite{xie2019improved}.

TabNet, a more recent addition to the fraud detection toolkit, is a deep learning architecture specifically designed for tabular data \cite{arik2021tabnet}. It utilizes an attentive mechanism to learn the importance of each feature and creates interpretable explanations for its predictions. TabNet has shown promising results in fraud detection, particularly in scenarios where interpretability is crucial\cite{huang2021tabnet,li2023mask}.

Despite the success of these techniques, they often face challenges when dealing with complex, high-dimensional data and long-range dependencies. This is where Transformer models, originally developed for natural language processing tasks, have shown immense potential\cite{chen2021pareto}. Haowei's work\cite{ni-24-time-series} has demonstrated the superiority of transformer-based models over traditional statistical approaches in accurately predicting heart rate time series, effectively capturing complex temporal dependencies and non-linear relationships inherent in physiological data. Despite the success of these techniques, they often face challenges when dealing with complex, high-dimensional data and long-range dependencies. This is where Transformer models, originally developed for natural language processing tasks, have shown immense potential\cite{chen2021pareto, zhang2024cunet}. Transformers can effectively capture intricate patterns and relationships within the data, thanks to their self-attention mechanism. By adapting Transformers for fraud detection, we can leverage their ability to handle complex data structures and create more accurate and robust models.

Moreover, Transformers can be pre-trained on large, unlabeled datasets and then fine-tuned for specific fraud detection tasks. This transfer learning approach allows the models to learn general representations of the data and reduces the need for extensive labeled datasets\cite{lyu2023attention}. Transformers have the potential to outperform traditional methods and provide a more efficient and effective solution for fraud detection in various domains. Therefore, in this study, we aim to thoroughly explore the feasibility of applying Transformers to fraud detection.

The remainder of this research is written  as follows. In Section II, we provide a comprehensive review of the related research that has informed the objectives and methodological choices of the current study. This review covers key contributions in areas such as statistical modeling, NN, and machine learning approaches to credit card fraud detection\cite{xu2024text}.

Section III then delves into the detailed methodology employed in this work, explaining the data processing, sampling strategies, model architectures, and underlying principles that constitute our fraud detection framework.

Moving to Section IV, we present the results of our extensive experimental evaluation, comparing the performance of our proposed model against several widely-used benchmark methods to rigorously assess the detection accuracy achieved.

Finally, Section V concludes the paper by summarizing the key findings, discussing the practical implications of our work, and outlining promising directions for future research in the field of credit card fraud detection.

\section{Related Work}
The detection of fraud in financial transactions has long been a pressing research challenge in the field of payment systems. As the use of mobile payments, digital wallets, and web-based payment methods has proliferated, accurately identifying the small proportion of fraudulent transactions amidst the vast volume of payment data has become an increasingly critical issue.

The academic literature has accumulated a rich body of research addressing this problem. Bolton and Hand\cite{bolton2002statistical} pioneered the use of statistical models to analyze and detect the presence of fraudulent activity. Building upon this foundation, Stojanović\cite{stojanovic2021follow} subsequently employed more advanced neural network models for fraud detection. Christoph\cite{DBLP:journals/pami/LampertNH14} proposed a machine learning approach involving feature vectorization and classification, providing an experimental framework for related studies. Furthermore, the work of Weng et al\cite{Weng2024}. offered valuable insights by demonstrating how data analysis and AI techniques can be leveraged to tackle data security and cybersecurity challenges, inspiring the focus of the current research.

Similar work has been done in other deep learning biological areas such as \cite{lai2023detect, jin2024apeer} fine-tuning segment anything to obtain accurate forged regions \cite{lai2024selective} proposing a new style sampling module that improves the detection accuracy and generalization of the anomalous face classification task

In the realm of model selection, the success of large language models in other domains, as showcased by the research of Wang et al\cite{wang2024enhancing, lyu2023backdoor}., has provided important guidance. To ensure the efficacy of training and testing strategies, the data cleaning and processing methods developed by Wei et al\cite{wei2023intags}. have been invaluable. Shen et al\cite{Shen2024}s work on efficient classification models has offered valuable experience in model selection, while Ding et al.'s\cite{ding-24-style} approach to data filtering, segmentation, and augmentation has provided useful insights for addressing sparse data challenges.

In their groundbreaking paper, Ni et al. (2023) present a comprehensive comparison between traditional models and deep learning approaches for heart rate time series forecasting\cite{ni-24-time-series}.Their research demonstrates that transformer-based models, particularly PatchTST, significantly outperform traditional models like ARIMA and Prophet in predicting heart rate dynamics\cite{ni-24-time-series}.

When it comes to experimental comparisons, the detailed evaluations of supervised learning, unsupervised learning, and deep learning methods conducted by Kazeem et al\cite{kazeemfraud}. have served as a useful reference. Additionally, the rigorous experimental and analytical methodologies employed by Jha et al\cite{jha2009credit}. have informed the meticulous implementation of the current study.

This rich body of prior research has provided a solid foundation and has inspired novel ideas for our own exploration. Building upon these insights, we aim to contribute innovative methods that advance the state of the art in credit card fraud detection, offering more effective solutions for the industry.

\section{Methodology}
In Section III, we provide a detailed account of the methodology employed in this work. This includes a thorough discussion of the data processing techniques utilized, such as ensuring an equally distributed dataset. We also examine the relevant prior research on feature engineering and dimensionality reduction, and analyze the impact these techniques had on the data used in our study\cite{zhao2024utilizing, yang_zhao_gao_2024}. By comprehensively documenting these methodological components, we aim to establish transparency and enable the replication of our research approach.

The primary focus of our methodology lies in accurately identifying instances of credit card fraud in datasets where fraudulent samples are exceedingly rare. We conducted comprehensive evaluations and explorations of various model architectures\cite{zhao_gao_yang_2024}, accompanied by detailed and thorough data analysis and preprocessing.\cite{Weng2024, jin2024learning} Additionally, we selected relevant loss functions and evaluation methods to ensure the robustness and effectiveness of our approach. The study\cite{ni-24-time-series} has significantly influenced our research, guiding the development of advanced predictive models in our work to handle complex temporal dependencies and non-linear relationships.

\subsection{Dataset Introduction}

The dataset used in this study consists of European credit card transaction data. To ensure the timeliness of the experiments and adapt to the latest payment environment, we utilized the most recent data available, encompassing over 550,000 transaction records of European credit card users up to 2023. These data contain the latest processed variables, ranging from V1 to V28. To enhance the informative value of the data and increase the scientific validity of the comparisons, we additionally incorporated European payment data from 2013, thereby increasing the datasource size and improving the meaningfulness of our experimental comparisons. Among the dataset’s 284,804 transactions, 492 were identified as fraudulent, accounting for 0.172\% of the total. All these data originate from the same source and possess variables V1 to V28, which have been processed and converted to float values. 

\subsection{Data Processing}

We first resample and preprocess the dataset to address the issue of data imbalance. By randomly sampling and merging, a new balanced dataset is generated, containing an equal number of fraudulent and non-fraudulent transaction samples. The new data set is then shuffled to ensure the randomness and reliability of the samples.

\subsubsection{Equally Distributing and Correlating}
To generate an equally distributed dataset, the initial step involves displaying textual output showcasing the proportions of class distribution to confirm the balance of the dataset Following this, a bar plot is generated utilizing Seaborn's countplot function to visually depict the distribution of classes. This visualization illustrates an equal distribution of classes, thereby underscoring the effectiveness of the subsampling technique in achieving class balance.

\subsubsection{Feature Correlation Analysi}
\begin{figure}
    \centering
    \includegraphics[width=1.0\linewidth]{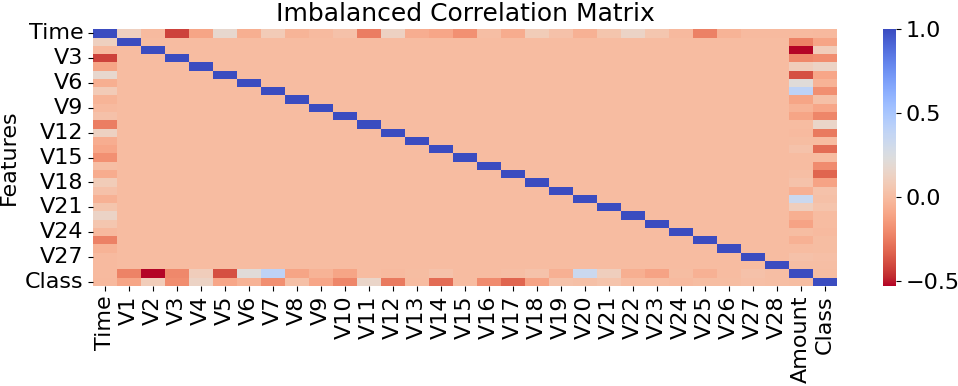}
    \caption{Imbalanced Correlation matrix.}
    \label{fig:corelation}
\end{figure}

\begin{figure}
    \centering
    \includegraphics[width=1\linewidth]{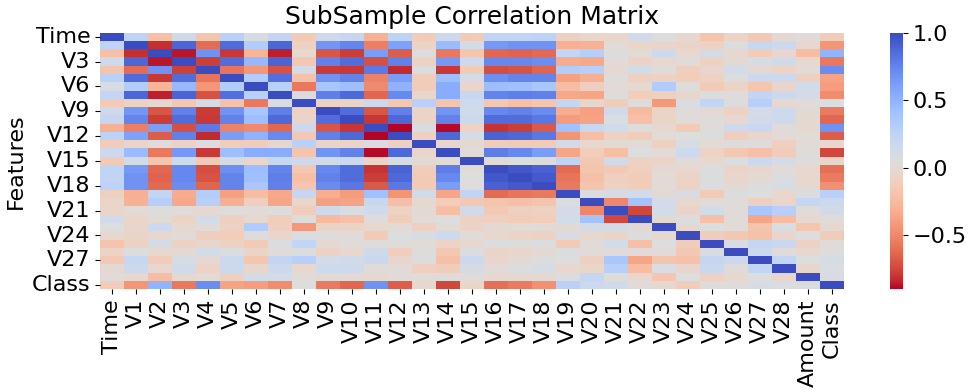}
    \caption{Sub-sampled correlation matrix.}
    \label{fig:balance-relation}
\end{figure}

In our analysis, we meticulously compared the original imbalanced dataset with its subsampled counterpart, unearthing a plethora of advantages stemming from data balancing. This encompassed a noticeable boost in model performance, mitigated susceptibility to overfitting, heightened predictive stability, and augmented model interpretability. The correlation efficacy visibly escalated as evidenced by the comparison, underscoring the tangible benefits derived from this approach. Correspond info please check Fig\ref{fig:corelation} and Fig\ref{fig:balance-relation}.

\subsubsection{Outlier Detection}
\begin{figure}
    \centering
    \includegraphics[width=1\linewidth]{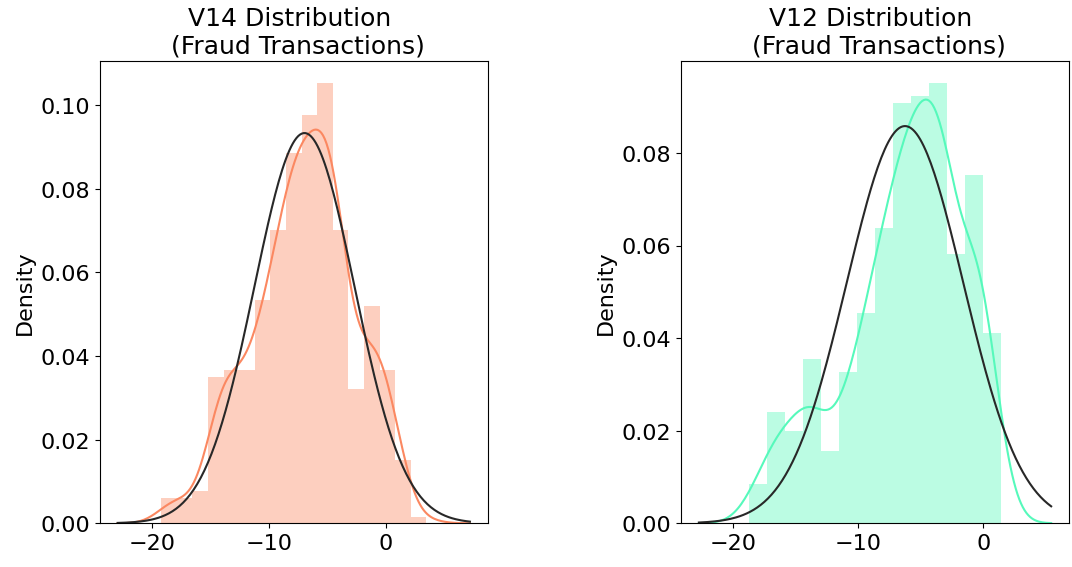}
    \caption{Outlier Detection for V14 and V12 feature.}
    \label{fig:out-lier-before}
\end{figure}

To better understand and handle outliers in the data, we first conducted a visual analysis of the data. Specifically, for features V14, V12, and V10, we plotted the distribution of these features in fraudulent transactions and fitted a normal distribution curve. Using the seaborn and matplotlib libraries, we created three side-by-side subplots, each showing the distribution of a single feature. By observing the distribution plots, we were able to preliminarily identify outliers in the data. This step of visual analysis helps to intuitively understand how these features manifest in fraudulent transactions as Fig\ref{fig:out-lier-before}. 

\begin{figure}
    \centering
    \includegraphics[width=0.9\linewidth]{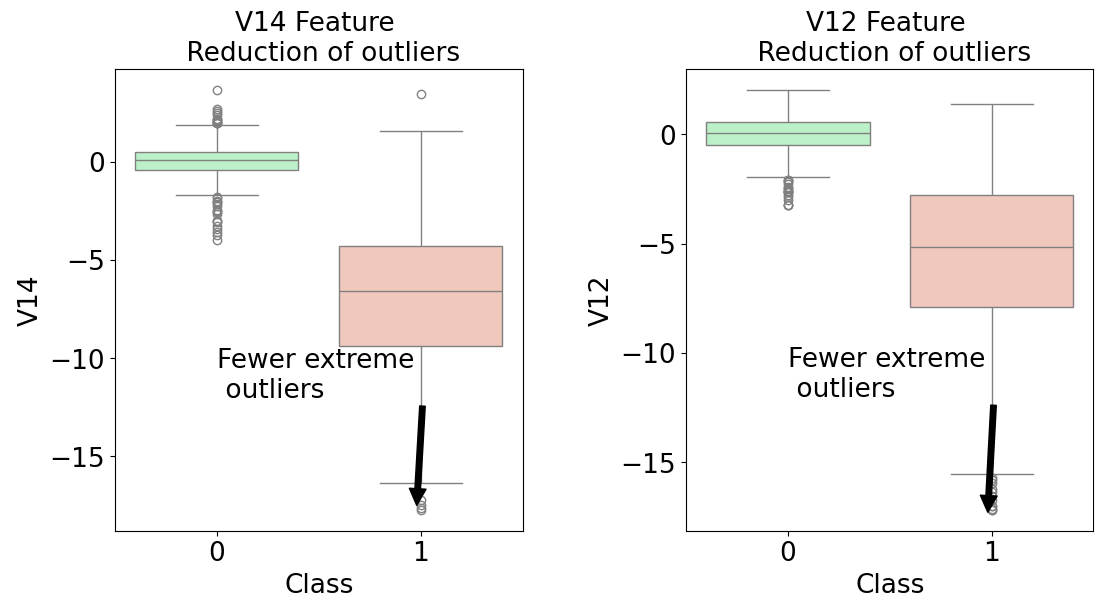}
    \caption{Outlier Removal for V14 and V12 feature.}
    \label{fig:outliearl}
\end{figure}

After identifying potential outliers, we formally handled these outliers using the Interquartile Range (IQR) method. The specific steps are as follows:

Calculate quartiles and interquartile range: First, calculate the 25th percentile (Q1) and 75th percentile (Q3) of the feature values. Then, calculate the interquartile range (IQR = Q3 - Q1).
Determine the upper and lower bounds for outliers: Use 1.5 times the IQR to determine the upper and lower bounds for outliers, i.e., the lower bound is Q1 - 1.5 * IQR, and the upper bound is Q3 + 1.5 * IQR.
Remove outliers: Consider feature values that are lower than the lower bound or higher than the upper bound as outliers and remove these outliers from the dataset. For example, for feature V14, we first calculate its quartiles and interquartile range, then determine the upper and lower bounds for outliers, and finally remove outliers that exceed these ranges. The same method is also applied to features V12 and V10.
To verify the effect of outlier handling, we again conducted a visual analysis of the processed data, using box plots to show the distribution of features V14, V12, and V10 in fraudulent and non-fraudulent transactions. Box plots not only display the central tendency and dispersion of the data but also visually show outliers as Fig \ref{fig:outliearl}.

By comparing the distribution plots before and after processing, we can observe the reduction of outliers and their impact on the data distribution. This step further verifies the effectiveness of our outlier handling method and provides a cleaner and more reliable dataset for subsequent model training

\subsubsection{Dimensionality Reduction}
To achieve optimal dimensionality reduction for the data, three different techniques—T-SNE, PCA, and Truncated SVD—are employed to map the data onto a two-dimensional space. This facilitates a visual representation of the distribution of data points on scatter plots, aiding in a better understanding of the data's structure and clustering tendencies.

\paragraph{T-SNE}
T-Distributed Stochastic Neighbor Embedding, is a technique for reducing the dimensionality of high-dimensional data, particularly when the goal is to represent the data in a low-dimensional space of two or three dimensions. Unlike linear dimensionality reduction methods, T-SNE is capable of capturing and preserving the complex nonlinear structures and local relationships inherent in high-dimensional data. The algorithm works by representing each high-dimensional data point as a corresponding low-dimensional point in a way that ensures similar objects are represented by nearby points and dissimilar objects are represented by distant points with a high degree of probability. This unique ability to maintain the intricate patterns and local similarities of the original data makes T-SNE an invaluable tool for data visualization and clustering analysis, providing researchers with a clear and intuitive understanding of the underlying structure and relationships within the high-dimensional data. 

\paragraph{PCA}
PCA (Principal Component Analysis) is a linear dimensionality reduction technique that transforms the original data into a new coordinate system where the greatest variances by any projection of the data come to lie on the first coordinates (called principal components), the second greatest variances on the second coordinates, and so on. PCA is efficient and widely used for reducing the dimensionality of datasets while preserving as much variability as possible.

\paragraph{Truncated SVD}
Truncated SVD (Singular Value Decomposition) is technique that decomposes a matrix into three other matrices. It is particularly useful for sparse data and is often used in the context of Latent Semantic Analysis (LSA) in natural language processing. Unlike PCA, Truncated SVD does not center the data before performing the decomposition, making it suitable for term-document matrices and other non-centered data.

\begin{figure}
    \centering
    \includegraphics[width=0.9\linewidth]{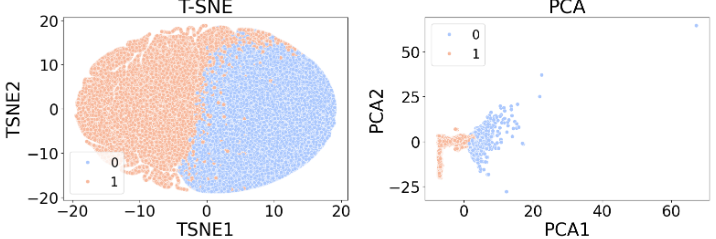}
    \caption{Comparison of T-SNE, PCA, and Truncated SVD. }
    \label{fig:comparison}
\end{figure}

\begin{figure}
    \centering
    \includegraphics[width=0.45\linewidth]{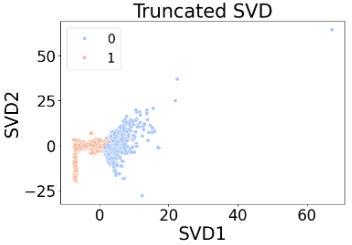}
    \caption{Detailed view of SVD results.}
    \label{fig:svdl}
\end{figure}

Compared to PCA and Truncated SVD, T-SNE excels in preserving the nonlinear structures and local relationships within high-dimensional data. Consequently, T-SNE offers superior advantages in data visualization and clustering analysis. The specific comparison results are illustrated in Figure \ref{fig:comparison} and Figure \ref{fig:svdl}.

\subsection{Model architectures}
\subsubsection{Transformer Structure}
The Transformer architecture employed in this study primarily consists of a Self-Attention Mechanism and a Feed-Forward Neural Network.

\paragraph{Self-Attention Mechanism}
The Self-Attention Mechanism is utilized to compute attention weights for each position in the input sequence, capturing dependencies between different positions. In the Transformer Encoder, the Self-Attention Mechanism is referred to as Multi-Head Attention. Below is the mathematical formulation of Multi-Head Attention:

Given an input sequence \( X = \{x_1, x_2, \ldots, x_n\} \), we first map the input sequence to Query (\(Q\)), Key (\(K\)), and Value (\(V\)) spaces. By matrix multiplication and scaling, we obtain Query (\(Q\)), Key (\(K\)), and Value (\(V\)) tensors:

\[
Q = XW_Q \quad K = XW_K \quad V = XW_V
\]

where \(W_Q\), \(W_K\), and \(W_V\) are learnable weight matrices.

Next, we compute the attention weights (\(A\)) by taking the dot product of Query and Key\cite{sun2022hierarchical}, followed by scaling and softmax normalization:

\[
A = \text{softmax}\left(\frac{QK^T}{\sqrt{d_k}}\right)
\]

Where \(d_k\) is the dimensionality of Query and Key.

Finally, we multiply the attention weights with the values and then perform a weighted sum to obtain the output of the Self-Attention Mechanism:

\[
Attention(X) = AV
\]

The Multi-Head Attention mechanism calculates the outputs of multiple attention heads by using independent Query, Key, and Value projection matrices. These outputs are concatenated along the last dimension to generate the final result of the Self-Attention Mechanism.

\paragraph{Operations in Transformer Encoder Layer}
In the Transformer encoder layer, we apply the self-attention mechanism to the input sequence, followed by a fully connected feed-forward neural network.

In the self-attention mechanism, we use the input sequence as queries, keys, and values and compute attention weights for each position with respect to other positions. Then, these attention weights are used to weight-sum the values of the input sequence, resulting in the output of the self-attention mechanism.

Next, the output of the self-attention mechanism is transformed through a feed-forward neural network, and the result is added back to the input (using residual connections), followed by layer normalization. Mathematically, it can be represented as:

\[
\begin{aligned}
\text{EncoderLayer}(X) &= \text{LayerNorm}(X + \text{Attention}(X)) \\
&+ \text{FeedForward}(X)
\end{aligned}
\]

Here, \(X\) represents the input sequence, \(\text{Attention}\) represents the self-attention mechanism, and \(\text{FeedForward}\) represents the feed-forward neural network.

Here is the Structure of our implemented model. Figure 6:
\begin{figure}
    \centering
    \includegraphics[width=1\linewidth]{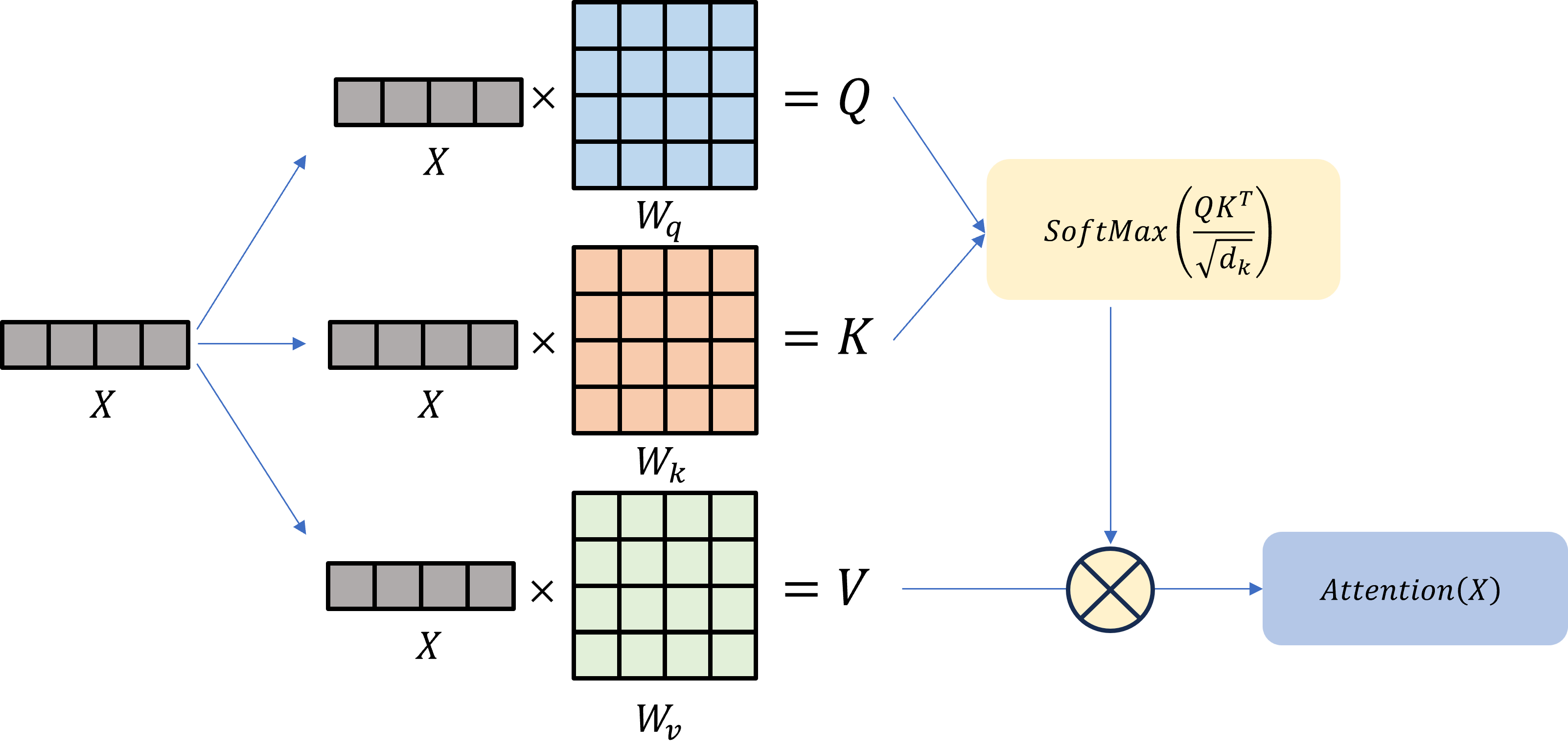}
    \caption{Structure of attention module.}
    \label{fig:enter-label}
\end{figure}

\section{Evaluation}
In this section, we first introduce the evaluation metrics, and then present our experiment results.

\subsection{Evaluation Metrics}
We use Precision, Recall, F1-score, and ROC AUC to assess our classification model. These metrics offer a thorough evaluation from different perspectives.

\paragraph{Precision}
Precision calculates the rate of true positive predictions among all positive predictions, indicating the model's accuracy in identifying positives\cite{gao2023hybrid}.

\paragraph{Recall}
Recall, or sensitivity, calculates the proportion of true positives detected among all actual positives, showing the model's ability to capture positive cases.

\paragraph{F1-score}
The F1-score provides a balanced measure of the model's accuracy in predicting positives and capturing actual positives.

\paragraph{ROC AUC}
ROC AUC evaluates the model's ability to distinguish between classes at various thresholds. It represents the area under the ROC curve, where a higher AUC indicates better discrimination between positive and negative instances.

The choice of metrics depends on the problem's requirements, with some scenarios prioritizing recall and others precision. The F1-score and ROC AUC offer balanced evaluations, considering both aspects.

\paragraph{F1-score}
F1-score is the harmonic mean of precision and recall, providing a balanced measure that combines both metrics' performance. It can be represented by the following formula:

\[
\text{F1-score} = \frac{2 \cdot \text{Precision} \cdot \text{Recall}}{\text{Precision} + \text{Recall}}
\]

\subsection{Results}
To fairly and accurately select the best fraud detection model, we carefully curated several different classification models. The models designed for this experiment include Logistic Regression, K-Nearest Neighbors, SVC, Decision Tree Classifier, and Neural Network. Additionally, to ensure that the models used in our experiments remain highly practical in comparison to the currently used models in the same scenarios, we included comparisons with the well-known and excellent model XGBoost and the deep learning-based TabNet.

Our experiments are divided into two stages. First, we used the latest data from 2023 to fairly compare the overall performance of all models, including Recall, Precision, F1 Score, and ROC AUC. To further validate the effectiveness of our models, we used the same models and processing methods to compare with the European fraud data covering both 2023 and 2013. Our aim is to verify whether our new models have an advantage over classic and well-known modern models, and whether this advantage is up-to-date and broadly applicable through repeated comparisons.

\begin{table}[h!]
\centering
\caption{Experiment with 2023 Data.}
\begin{tabular}{|l|c|c|c|c|}
\hline
\textbf{Classifier}       & \textbf{Precision} & \textbf{Recall} & \textbf{F1 Score} & \textbf{ROC AUC} \\ \hline
Logistic Regression       & 0.93               & 0.93            & 0.93              & 0.98             \\ \hline
KNN                       & 0.93               & 0.93            & 0.93              & 0.98             \\ \hline
SVM                       & 0.91               & 0.91            & 0.91              & 0.99             \\ \hline
Decision Tree             & 0.93               & 0.93            & 0.93              & 0.93             \\ \hline
Neural Network            & 0.92               & 0.91            & 0.91              & 0.96             \\ \hline
XGBoost                   & 0.95               & 0.95            & 0.95              & 0.99             \\ \hline
TabNet                    & 0.93               & 0.93            & 0.93              & 0.98             \\ \hline
Transformer               & 0.998              & 0.998           & 0.998             & 0.99             \\ \hline
\end{tabular}
\label{table:performance}
\end{table}

\begin{figure}[h!]
\centering
\includegraphics[width=0.9\linewidth]{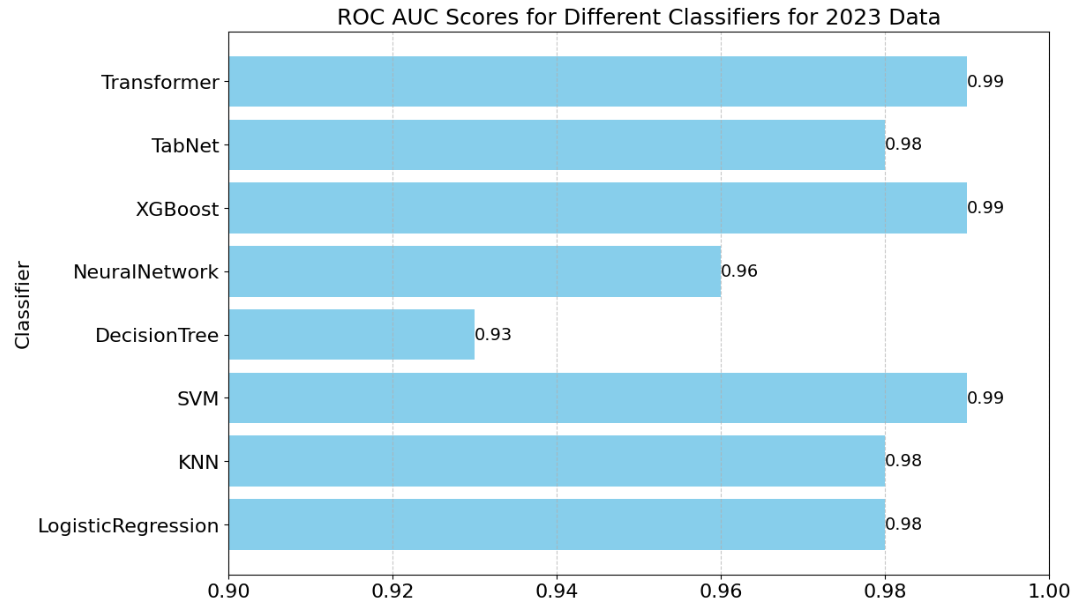} 
\caption{ROC and AUC Scores of Various Classifiers on 2023 Data. }
\label{fig:fig7}
\end{figure}

As shown in Table \ref{table:performance} and Fig \ref{fig:fig7}, the Transformer model has demonstrated remarkable performance in this experiment, outperforming other models such as XGBoost and TabNet across various evaluation metrics. Firstly, the Transformer achieves significantly higher Precision and Recall scores of 0.998, surpassing XGBoost (0.95), TabNet (0.93), and Neural Network (0.92 and 0.91) \cite{vaswani2017attention}. This indicates that the Transformer excels at correctly identifying positive samples while minimizing false positives and false negatives.

Moreover, the Transformer's F1 Score, which balances Precision and Recall, reaches an impressive 0.998, considerably higher than XGBoost (0.95), TabNet (0.93), and Neural Network (0.91) \cite{devlin2018bert}. This showcases the Transformer's ability to maintain a perfect equilibrium between Precision and Recall, resulting in superior overall performance.

To further validate the superior performance of the Transformer model in the domain of fraud detection, we conducted a cross-validation using data from 2013. The results demonstrate that the Transformer model consistently outperforms other models on the 2013 data, achieving significantly higher scores in Precision, Recall, F1-score, and ROC AUC.

\begin{table}[h!]
\centering
\caption{Performance Metrics of Various Classifiers on 2013 Data}
\begin{tabular}{|l|c|c|c|c|}
\hline
\textbf{Model}            & \textbf{Precision} & \textbf{Recall} & \textbf{F1-score} & \textbf{ROC AUC} \\ \hline
KNN                       & 0.93               & 0.92            & 0.92              & 0.95             \\ \hline
SVM                       & 0.93               & 0.93            & 0.93              & 0.95             \\ \hline
Decision Tree             & 0.91               & 0.90            & 0.89              & 0.85             \\ \hline
Logistic Regression       & 0.96               & 0.96            & 0.96              & 0.95             \\ \hline
Neural Network            & 0.975              & 0.999           & 0.988             & 0.85             \\ \hline
XGBoost                   & 0.96               & 0.90            & 0.92              & 0.96             \\ \hline
TabNet                    & 0.59               & 0.77            & 0.67              & 0.48             \\ \hline
Transformer               & 0.998              & 0.998           & 0.998             & 0.98             \\ \hline
\end{tabular}
\label{table:2013_performance}
\end{table}

\begin{figure}[h!]
    \centering
    \includegraphics[width=0.9\linewidth]{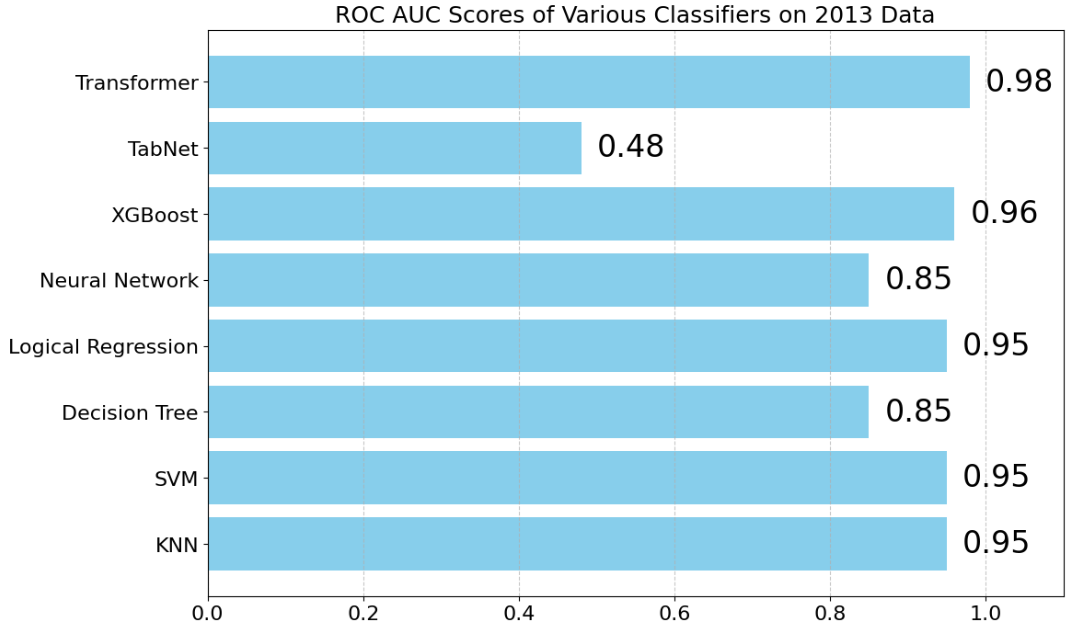}
    \caption{ROC AUC Scores of Various Classifiers on 2013 Data.}
    \label{fig:roc_auc_plot}
\end{figure}

Firstly, the Transformer model attains a Precision and Recall of 0.998 on the 2013 data, which is consistent with its performance on the 2023 data. This indicates that the Transformer model maintains a high level of accuracy and recall across different time periods, proving its excellent generalization ability and stability \cite{vaswani2017attention}. In contrast, although the Neural Network achieves a high Recall (0.999) on the 2013 data, its Precision (0.975) is slightly lower than that of the Transformer, suggesting that it is less effective in reducing false positives.

Secondly, the Transformer model's F1-score on the 2013 data also reaches 0.998, significantly surpassing other models. This further confirms the Transformer's superiority in balancing Precision and Recall . It is worth noting that TabNet performs poorly on the 2013 data, with an F1-score of only 0.67. This may be attributed to TabNet's sensitivity to changes in data distribution, resulting in unstable performance across different time periods.

Moreover, the Transformer model achieves an ROC AUC of 0.98 on the 2013 data, which is only slightly lower than XGBoost's 0.96. However, considering its performance in other metrics, the Transformer still demonstrates the best overall performance. This further validates the Transformer's advantage in capturing complex feature interactions and long-range dependencies, enabling it to maintain consistently high performance on data from different time periods .

In conclusion, the cross-validation using 2013 data further confirms the superior performance of the Transformer model in the field of fraud detection. Its outstanding Precision, Recall, F1-score, and ROC AUC, as well as its stable performance across different time periods, can be attributed to its unique self-attention mechanism and pretraining-fine-tuning paradigm. This makes the Transformer model an ideal choice for addressing complex fraud detection problems and provides strong support for its application in other domains.

\section{Conclusion}
In this experiment, we successfully adapted and optimized the Transformer model for fraud detection, achieving remarkable results. The experimental outcomes demonstrate that our Transformer model significantly outperforms other classical models, such as XGBoost, TabNet, and Neural Network, across multiple evaluation metrics, including Precision, Recall, F1-score, and ROC AUC. By conducting cross-validation on data from two different time periods, 2013 and 2023, we further confirmed the Transformer model's exceptional generalization ability and stability.

The significance of these results is twofold. First, it validates that Transformer models not only excel in general artificial intelligence applications like ChatGPT but also possess tremendous potential in classification tasks, surpassing traditional, classical approaches. Second, our innovative attempt to apply the Transformer in fraud detection lays a solid foundation for the future development of more optimized Transformer-based models, promising to enhance financial security measures further.

The success of our experiment can be attributed to our extensive efforts in understanding the underlying principles of the Transformer and adjusting its Encoder layers. Despite the considerable challenge of adapting the Transformer, initially designed for natural language processing tasks, into a classification model suitable for fraud detection, our model ultimately lived up to expectations, demonstrating outstanding accuracy across all evaluation metrics. This result not only proves the superior performance of the Transformer model in the fraud detection domain but also provides strong support for its application in other fields.

In the future research, our focus will remain on investigating the potential of Transformer models in fraud detection and beyond\cite{202407.0981, 10233897}. Our goal is to create models that are not only more efficient and accurate but also more resilient to various challenges. Moreover, we believe that the findings of this experiment will serve as a valuable resource for professionals in the industry, encouraging the adoption and advancement of AI-driven solutions in the realm of financial security and other critical areas.

\bibliographystyle{plain}
\bibliography{ref}

\end{document}